\documentclass[5p]{elsarticle}

\usepackage{algorithm}
\usepackage{algorithmic}
\usepackage{amsmath,amssymb,amsfonts}
\usepackage{balance}
\usepackage{bm}			% bold type for equations
\usepackage{booktabs}
\usepackage{color}		% color content
\usepackage{diagbox}
\usepackage{graphicx}
\graphicspath{{figure/}}
\DeclareGraphicsExtensions{.eps, .pdf}
\usepackage[caption=false,font=footnotesize,labelfont=sf,textfont=sf]{subfig}
\usepackage{lineno}
\usepackage{multirow}
\usepackage{newtxtext}
\usepackage{picinpar}
\usepackage{upgreek}
\usepackage{url}		% hyperlinks
\usepackage[colorlinks, bookmarks, unicode]{hyperref}

\modulolinenumbers[5]
\journal{}

\newcommand{\trace}[1]{\mathrm{tr}({#1})}
\newcommand{\norm}[1]{||{#1}||}
\newcommand{\transpose}{{\mathrm{T}}}
\newcommand{\fro}{\mathrm{F}}
\newcommand{\tensor}[1]{\bm{\mathcal{#1}}}
\newcommand{\ftensor}[1]{\bm{\mathcal{\bar{#1}}}}
\newcommand{\fmat}[1]{\bm{\bar{#1}}}
\newcommand{\Rdim}[1]{\in\mathbb{R}^{#1}}

\newcommand{\etal}{\emph{et al}.\ }

\newtheorem{definition}{Definition}[section]
\newtheorem{theorem}[definition]{Theorem}

\begin{document}

\begin{frontmatter}

\title{Truncated nuclear norm regularization for low-rank tensor completion}

\author{Shengke~Xue\corref{corr1}}
\cortext[corr1]{Corresponding author}
\ead{xueshengke@zju.edu.cn}

\author{Wenyuan~Qiu\corref{}}
\ead{qiuwenyuan@zju.edu.cn}

\author{Fan~Liu}
\ead{flyingliufan@zju.edu.cn}

\author{Xinyu~Jin}
\ead{jinxinyuzju@gmail.com}

\address{College of Information Science and Electronic Engineering, Zhejiang University, \\ No.~38 Zheda Road, Hangzhou~310027, China}

\begin{abstract}
Recently, low-rank tensor completion has become increasingly attractive in recovering incomplete visual data. Considering a color image or video as a three-dimensional (3D) tensor, existing studies have put forward several definitions of tensor nuclear norm. 
However, they are limited and may not accurately approximate the real rank of a tensor, and they do not explicitly use the low-rank property in optimization. It is proved that the recently proposed truncated nuclear norm (TNN) can replace the traditional nuclear norm, as an improved approximation to the rank of a matrix. 
In this paper, we propose a new method called the tensor truncated nuclear norm (T-TNN), which suggests a new definition of tensor nuclear norm. The truncated nuclear norm is generalized from the matrix case to the tensor case. With the help of the low rankness of TNN, our approach improves the efficacy of tensor completion. We adopt the definition of the previously proposed tensor singular value decomposition, the alternating direction method of multipliers, and the accelerated proximal gradient line search method in our algorithm. 
Substantial experiments on real-world videos and images illustrate that the performance of our approach is better than those of previous methods. 
\end{abstract}


\end{frontmatter}

%\linenumbers

\section{Introduction}

Recovering missing elements in high dimensional data has  gained cumulative attention in computer vision and pattern recognition. Discovering the inherent low-rank nature of incomplete data with partial observed elements has been widely studied in various applications, e.g., motion capture \cite{Hu2017-Motion}, face recognition \cite{Gao2017-Learning, Yang2017-Nuclear}, image alignment \cite{Song2016-Image, Peng2012-RASL}, object detection \cite{Wang2017-Salient, Zhang2017-Salient}, and image classification \cite{Li2014-Learning, Li2016-Learning, Xue2017-Robust}.

Estimating missing values in visual data is generally regarded as a low-rank matrix approximation problem, because it lies in a low dimensional space  \cite{Candes2009-Exact}.  The rank function, non-convex and NP-hard, is usually replaced by the nuclear norm. Existing studies indicate that the nuclear norm is appropriate to solve a large number of low-rank optimization problems. However, Hu \etal \cite{Hu2013-Accurate} declared that due to minimizing all of the singular values simultaneously, the nuclear norm may not approximate well to the rank function.

As a more accurate and much tighter alternative to the rank function, the truncated nuclear norm regularization (TNNR) \cite{Hu2013-Accurate} was proposed to replace the traditional nuclear norm. It is proved that the TNNR obviously improves the efficacy of image recovery. Specifically, the TNNR neglects the largest $r$ singular values of data and tries to optimize the smallest $\min(m,n)-r$ singular values, where $m$\,$\times$\,$n$ denotes the dimension of two-dimensional data and $r$ denotes the number of truncated values. Numerous studies were inspired by this. For example, Liu \etal \cite{Liu2016-Truncated} developed a weighted TNNR to further accelerate their algorithm, by using a gradient descent scheme; Lee and Lam \cite{Lee2016-Computationally} proposed the ghost-free high dynamic range imaging from irradiance maps by introducing the TNNR method. Combining with TNNR, Hu \etal \cite{Hu2015-Large} achieved large scale multi-class classification,  by using the lifted coordinate descent method. Lin \etal \cite{Lin2017-Factorization} applied TNNR to the factorization for projective and metric reconstruction. Hong \etal \cite{Hong2016-Online} proposed an  online robust principal component analysis algorithm by adopting the truncated nuclear norm.

In addition, most previous  low-rank matrix approximation methods  cope with the input data in a two-dimensional fashion. To be specific, the algorithms are employed on each channel individually and then the results are merged together, in the case of recovering a color image. It shows an explicit drawback that the structural information between channels is not involved. Thereby, recent studies consider a color image as 3D data and formulate it as a low-rank tensor completion problem.

As an  extension of the matrix case, tensor completion  becomes increasingly important. However, the definition of the nuclear norm of a tensor turns out to be difficult, since it cannot be intuitively derived from the matrix case. Several types of tensor nuclear norm have been proposed; however, they are pretty different from each other. Liu \etal \cite{Liu2013-Tensor} initially proposed the sum of matricized nuclear norms (SMNN) of a tensor, which is defined as follows:
\begin{equation}
\min_{\tensor{X}} \ \sum_{i=1}^k \alpha_i \norm{\tensor{X}_{[i]}}_* \ \ \text{s.t.} \ \ \tensor{X}_{\bm{\Omega}} = (\tensor{X}_0)_{\bm{\Omega}} , \label{eq:SNN}
\end{equation}
where $\tensor{X}_{[i]}$ denotes the matrix of the tensor unfolded along the $i$th dimension (i.e., the mode-$i$ matricization of $\tensor{X}$), $\alpha_i > 0$ is a parameter that satisfies $\sum_{i=1}^n \alpha_i = 1$, $\tensor{X}_0$ is the original incomplete data, and $\bm{\Omega}$ is the set of positions relating to known elements. Hosono \etal \cite{Hosono2016-Weighted} and Zhang \etal \cite{Zhang2018-Nonlocal} used the SMNN for nonlocal image denoising, both of which obtained visually and quantitatively improved results. But so far no theoretical analysis declares that the nuclear norm of each matricization of $\tensor{X}$ is plausible since the spacial structure may lose due to the matricization. Additionally, it is not clear  to decide the optimal value of $\alpha_i$ \cite{Zhang2017-Exact}, though they directly dominate the weights of $k$ norms in problem \eqref{eq:SNN}. In general, $\alpha_i$ is empirically determined in advance.

Kilmer \etal \cite{Misha2013-Third-Order} proposed a novel tensor decomposition scheme, called the tensor singular value decomposition (t-SVD). Based on the new definition of the tensor-tensor product, some properties of the t-SVD are quite similar to the matrix case.  Zhang \etal \cite{Zhang2014-Novel} declared their tubal nuclear norm (Tubal-NN) as the sum of nuclear norms of all frontal slices in the Fourier domain and clarified that it was a convex relaxation to the tensor rank. Their optimization problem can be formulated as
\begin{equation} \label{eq:SFNN}
\min_{\tensor{X}} \ \sum_{i=1}^{n_3} \norm{\fmat{X}^{(i)}}_* \ \ \text{s.t.} \ \ \tensor{X}_{\bm{\Omega}} = (\tensor{X}_0)_{\bm{\Omega}} , 
\end{equation}
where $\fmat{X}^{(i)}$ will be introduced in Section~\ref{sec:notations}. Semerci \etal \cite{Semerci2014-Tensor-Based} used this model to multienergy computed tomography images and achieved encouraging effects of reconstruction. Based on the Tubal-NN, Liu \etal \cite{Liu2016-Adaptive} considered the 3D radio frequency fingerprint data as tensors for fine-grained indoor localization. However, the low-rank property was not explicitly considered in optimization, and it still entailed a vast number of iterations to converge. Since the t-SVD is a sophisticated function, the overall computational cost of \eqref{eq:SFNN} will be highly expensive.

This study is an extension of our  conference paper \cite{Xue2018-Low-Rank}. In this paper, we propose a new approach called the tensor truncated nuclear norm (T-TNN). Based on the t-SVD, we define that our tensor nuclear norm is  the sum of all singular values in an f-diagonal tensor. This is extended directly from the matrix case. In addition, we validate that our T-TNN can be computed efficiently in the Fourier domain. 
To further take the advantage of TNNR, our T-TNN method generalizes it to the 3D case. Following common strategies, we adopt the universal alternating direction method of multipliers (ADMM) \cite{Lin2011-Linearized} and the accelerated proximal gradient line search method (APGL)  \cite{Beck2009-Iterative} to solve our optimization problem. Experimental results validate that our approach outperforms  previous methods.

%Due to the benefit from the TNN, the computational time and the required number of iterations are decreased substantially.

The remainder of this paper is organized as follows. Section~\ref{sec:notations} introduces some notations and definitions. Section~\ref{sec:tnnr} shows the entire framework of our T-TNN. In Section~\ref{sec:experiment}, experimental results evaluate the performance of our approach. Section~\ref{sec:conclusion} states the conclusions and our future work.

\section{Notations and preliminaries} \label{sec:notations}

%\begin{figure}
%	\centering
%	\includegraphics[scale=0.30]{tensor.pdf}
%	\caption{Illustration of a 3-order tensor}
%	\label{fig:tensor}
%\end{figure}

Some basic notations and definitions used in this paper are summarized in Table~\ref{tab:notations}.

\begin{table}
	\small \centering
	\caption{Notations and definitions} \label{tab:notations}
	\addtolength{\tabcolsep}{-3pt}
	\centering
	\begin{tabular}{cl}
		\toprule
		                 Symbol                  &                              \multicolumn{1}{c}{Description}                              \\ 
        \midrule
		 $\tensor{A} / \bm{A} / \bm{a} / a$   &                    Tensor / Matrix / Vector / Scalar                     \\ 
		               $\bm{I}_n$                &                            Identity matrix                            \\ 
%		      $\mathbb{R} / \mathbb{C}$        &                 Field of real numbers / complex                  \\ 
%		      & \quad numbers \\
		           $\tensor{A}_{ijk}$            &                  ($i,j,k$)th element of $\tensor{A}$                  \\ 
		           $\tensor{A}(i,:,:) / \tensor{A}(:,i,:)$ & $i$th horizontal / lateral slice of $\tensor{A}$ \\		           		       
		             $\tensor{A}(:,:,i) / \bm{A}^{(i)}$              &  $i$th frontal slice of $\tensor{A}$ \\ 		           
		$\trace{\cdot} / (\cdot)^\transpose$ &                  Trace function / Conjugate transpose                  \\ 
		    $\langle \bm{A}, \bm{B} \rangle \triangleq \trace{ \bm{A}^\transpose \bm{B}}$     &                             Inner product of matrices                            \\ 
		    $\langle \tensor{A}, \tensor{B} \rangle \triangleq \sum_{i=1}^{n_3} \langle \bm{A}^{(i)}, \bm{B}^{(i)} \rangle $ & Inner product of tensors \\
		    $\trace{\tensor{A}} = \sum_{i=1}^{n_3} \trace{\bm{A}^{(i)}}$ & Trace of a tensor \\
		    $\norm{\tensor{A}}_1 \triangleq \sum_{ijk} |\tensor{A}_{ijk}|$ & $\ell_1$ norm \\
		    $\norm{\tensor{A}}_\infty \triangleq \max_{ijk} |\tensor{A}_{ijk}|$		    & Infinity norm \\
		    $\norm{\tensor{A}}_\fro \triangleq \sqrt{\sum_{ijk}|\tensor{A}_{ijk}|^2}$		    		    & Frobenius norm \\
		    $\norm{\bm{A}}_* \triangleq \sum_{i} \sigma_i (\bm{A})$ & Matrix nuclear norm, i.e., sum of all  \\
		    & \quad singular values \\
		\bottomrule
	\end{tabular}
\end{table}

For tensor $\tensor{A} \Rdim{n_1 \times n_2 \times n_3}$, by using the Matlab notation, we define $\ftensor{A} \triangleq \textsf{fft}(\tensor{A},[\,],3)$, which is the discrete Fourier transform of $\tensor{A}$ along the third dimension. Similarly, we  compute $\tensor{A} \triangleq \textsf{ifft}(\ftensor{A},[\,],3)$ via the inverse $\textsf{fft}$ function. We define $\fmat{A}$ as a block diagonal matrix, where each frontal slice $\fmat{A}^{(i)}$ of $\ftensor{A}$ lies on the diagonal in order, i.e.,
\begin{equation}
\fmat{A} \triangleq \textsf{bdiag}(\ftensor{A}) \triangleq 
\begin{bmatrix}
\fmat{A}^{(1)} &                    &        &                      \\
& \fmat{A}^{(2)} &        &                      \\
&                    & \ddots &                      \\
&                    &        & \fmat{A}^{(n_3)}
\end{bmatrix} .
\end{equation}

The block circulant matrix of tensor $\tensor{A}$ is defined as
\begin{equation} \label{eq:bcirc}
\textsf{bcirc}(\tensor{A}) \triangleq 
\begin{bmatrix}
\bm{A}^{(1)}   & \bm{A}^{(n_3)}   & \cdots & \bm{A}^{(2)} \\
\bm{A}^{(2)}   & \bm{A}^{(1)}     & \cdots & \bm{A}^{(3)} \\
\vdots         & \vdots           & \ddots & \vdots       \\
\bm{A}^{(n_3)} & \bm{A}^{(n_3-1)} & \cdots & \bm{A}^{(1)}
\end{bmatrix} .
\end{equation}

Here, a pair of folding operators are defined as follows: 
\begin{equation} \label{eq:unfold}
\textsf{unfold}(\tensor{A}) \triangleq 
\begin{bmatrix}
\bm{A}^{(1)} \\
\bm{A}^{(2)} \\
\vdots \\
\bm{A}^{(n_3)} 
\end{bmatrix} \! , \
\textsf{fold}(\textsf{unfold}(\tensor{A})) \triangleq \tensor{A} \, .
\end{equation}

\begin{definition}[tensor product] \label{def:tensor_product} \emph{\cite{Misha2013-Third-Order}}
	With $\tensor{A} \Rdim{n_1 \times n_2 \times n_3}$ and $\tensor{B} \Rdim{n_2 \times n_4 \times n_3}$,  the tensor product $\tensor{A} * \tensor{B}$ is defined as a tensor with size $n_1 \times n_4 \times n_3$, i.e.,
	\begin{equation}
	\tensor{A} * \tensor{B} \triangleq \emph{\textsf{fold}} (\emph{\textsf{bcirc}} (\tensor{A}) \cdot \emph{\textsf{unfold}} (\tensor{B}) ) .
	\end{equation}
	The tensor product is similar to the matrix product except that the multiplication between elements is replaced by the circular convolution.
	Notice that the tensor product reduces to the standard matrix product if $n_3 = 1$.
\end{definition}

\begin{definition}[conjugate transpose] \emph{\cite{Misha2013-Third-Order}}
	Define the conjugate transpose of tensor $\tensor{A} \Rdim{n_1 \times n_2 \times n_3}$ as $\tensor{A}^{\transpose} \Rdim{n_2 \times n_1 \times n_3}$. It is obtained by conjugate transposing each frontal slice and then reversing the order of transposed frontal slices 2 to $n_3$:
	\begin{equation}
	\begin{aligned}
	\big( \tensor{A}^{\transpose} \big)^{(1)} & \triangleq \big( \tensor{A}^{(1)} \big)^\transpose, \\ 
	\big( \tensor{A}^{\transpose} \big)^{(i)} & \triangleq \big( \tensor{A}^{(n_3 + 2 - i)} \big)^\transpose, \ i = 2, \ldots, n_3.
	\end{aligned}
	\end{equation}
\end{definition}

\begin{definition}[identity tensor] \emph{\cite{Misha2013-Third-Order}}
	Define $\tensor{I} \Rdim{n \times n \times n_3}$ as an identity tensor,  whose first frontal slice $\bm{I}^{(1)}$ is an $n \times n$ identity matrix and the other slices are zero.
\end{definition}

\begin{definition}[orthogonal tensor] \emph{\cite{Misha2013-Third-Order}}
	The	orthogonal tensor $\tensor{Q} $ satisfies the following:
	\begin{equation}
	\tensor{Q} * \tensor{Q}^{\transpose} \triangleq \tensor{Q}^{\transpose} * \tensor{Q} \triangleq \tensor{I}.
	\end{equation}
\end{definition}

\begin{definition}[f-diagonal tensor] \emph{\cite{Misha2013-Third-Order}}
	Tensor $\tensor{A}$ is called f-diagonal if each frontal slice $\bm{A}^{(i)}$ is a diagonal matrix.
\end{definition}

\begin{figure}
	\centering \small
	\includegraphics[scale=0.28]{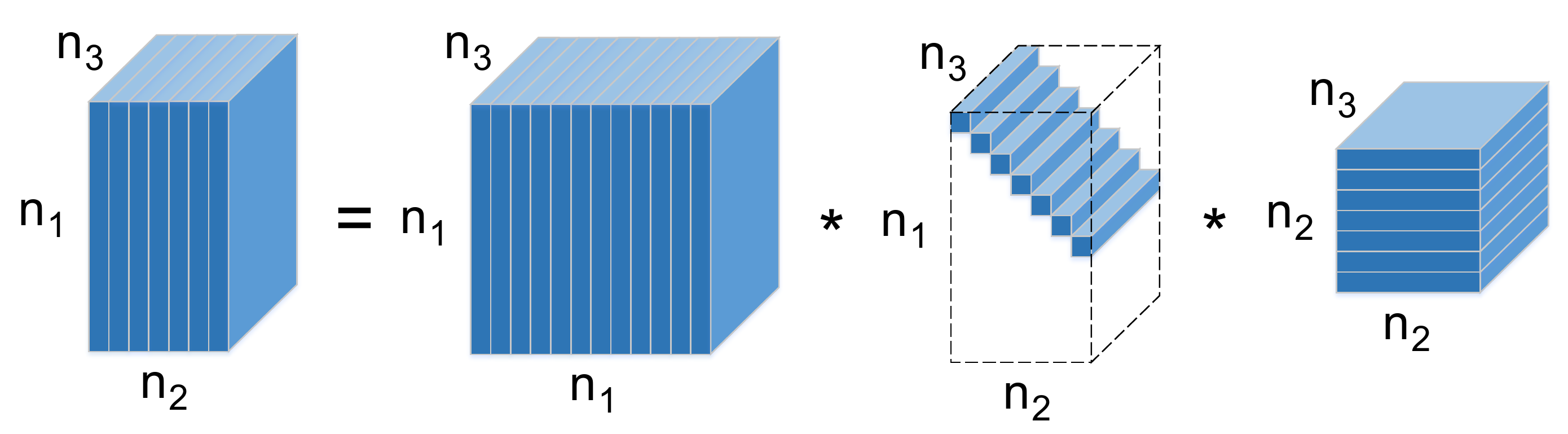}
	\caption{Illustration of the t-SVD of an $n_1$\,$\times$\,$n_2$\,$\times$\,$n_3$ tensor}
	\label{fig:t-SVD}
\end{figure}

\begin{theorem}[tensor singular value decomposition] \label{the:t-SVD} \emph{\cite{Misha2013-Third-Order}} \\
	Tensor $\tensor{A} \Rdim{n_1 \times n_2 \times n_3}$  can be decomposed as
	\begin{equation}
	\tensor{A} \triangleq \tensor{U} * \tensor{S} * \tensor{V}^{\transpose},
	\end{equation}
	where $\tensor{U} \Rdim{n_1 \times n_1 \times n_3}$ and $\tensor{V} \Rdim{n_2 \times n_2 \times n_3}$ are orthogonal, and $\tensor{S} \Rdim{n_1 \times n_2 \times n_3}$ is an f-diagonal tensor.
\end{theorem}
Fig.~\ref{fig:t-SVD} illustrates the t-SVD. It can be efficiently carried out based on the matrix SVD in the Fourier domain, because of  an important property that the block circulant matrix can be converted to a block diagonal matrix in the Fourier domain, i.e.,
\begin{equation}
(\bm{F}_{n_3} \otimes \bm{I}_{n_1}) \cdot \textsf{bcirc} (\tensor{A}) \cdot (\bm{F}_{n_3}^\transpose \otimes \bm{I}_{n_2}) = \fmat{A},
\end{equation}
where $\bm{F}_{n_3}$ denotes the $n_3 \times n_3$ discrete Fourier transform matrix and $\otimes$ denotes the Kronecker product. Note that the matrix SVD can be performed on each frontal slice of $\ftensor{A}$, i.e., $\fmat{A}^{(i)} = \fmat{U}^{(i)} \fmat{S}^{(i)} \fmat{V}^{(i)\transpose}$, where  $\fmat{U}^{(i)}$, $\fmat{S}^{(i)}$, and $\fmat{V}^{(i)}$ are the frontal slices of  $\,\ftensor{U}$, $\ftensor{S}$, and $\ftensor{V}$, respectively. In brief, we have $\fmat{A} = \fmat{U} \fmat{S} \fmat{V}^\transpose$. By using the {\sf ifft} function along the third dimension, we obtain $\tensor{U} = \textsf{ifft}(\ftensor{U},[\,],3)$, $\tensor{S} = \textsf{ifft}(\ftensor{S},[\,],3)$, and $\tensor{V} = \textsf{ifft}(\ftensor{V},[\,],3)$.

\begin{definition}[tensor tubal rank and tensor nuclear norm] \label{def:tnn} 
	Let  the t-SVD of tensor $\tensor{A} \Rdim{n_1 \times n_2 \times n_3}$  be $\tensor{U} * \tensor{S} * \tensor{V}^{\transpose}$. The tensor tubal rank of $\tensor{A}$ is defined as the maximum rank among all frontal slices of the f-diagonal $\tensor{S}$, i.e., $\max_{i}\,\text{rank}(\bm{S}^{(i)})$. Our tensor nuclear norm $\norm{\tensor{A}}_*$ is defined as the sum of the singular values in all frontal slices of  $\tensor{S}$, i.e.,
	\begin{equation}
	\norm{\tensor{A}}_* \triangleq \trace{\tensor{S}} = \sum_{i=1}^{n_3} \trace{\bm{S}^{(i)}} .
	\end{equation} 
\end{definition}
Note that our tensor nuclear norm simplifies to standard matrix nuclear norm if $n_3 = 1$. Therefore, our tensor nuclear norm can be considered as a direct extension from the matrix case to the tensor case.

Because of the \textsf{fft} function in the third dimension, we exploit the symmetric property that the trace of tensor product $\tensor{A} * \tensor{B}$ is equal to the trace of the product of $\fmat{A}^{(1)}$ and $\fmat{B}^{(1)}$, which are the first frontal slices of  $\ftensor{A}$ and $\ftensor{B}$ in the Fourier domain, i.e.,
\begin{equation}
\trace{ \tensor{A} * \tensor{B} } = \trace{ \fmat{A}^{(1)} \fmat{B}^{(1)} } . \label{eq:fft_property}
\end{equation}
The proof is provided in Appendix. Derived from \eqref{eq:fft_property}, we can further simplify our tensor nuclear norm as follows:
\begin{equation}
\norm{\tensor{A}}_*  \triangleq \trace{\tensor{S}} =  \trace{\fmat{S}^{(1)}} = \norm{\fmat{A}^{(1)}}_* .
\end{equation}
This indicates that our tensor nuclear norm can be efficiently calculated by one matrix SVD in the Fourier domain, rather than through the complicated t-SVD to obtain $\tensor{S}$.

Our definition is different from those of previous studies \cite{Zhang2017-Exact, Lu2016-Tensor}, which are also defined in the Fourier domain. The tubal nuclear norm in \cite{Zhang2017-Exact} required to  compute each frontal slice of $\ftensor{S}$.  Similarly, Lu \etal \cite{Lu2016-Tensor} further suggested that their tensor  nuclear norm, which was used for robust principal component analysis (RPCA),  was equal to  the nuclear norm of the block circulant matrix of a tensor with factor $1/n_3$, i.e., $\norm{\tensor{A}}_* = \frac{1}{n_3} \norm{\textsf{bcirc}(\tensor{A})}_*$. However, $\textsf{bcirc}(\tensor{A}) \Rdim{n_1 n_3 \times n_2 n_3}$ requires a vast amount of memory if $n_1$, $n_2$, or $n_3$ is large. This makes the matrix SVD much slower.

\begin{definition}[singular value thresholding] \label{def:SVT}
	Assume that the t-SVD of tensor $\tensor{X} \Rdim{n_1 \times n_2 \times n_3}$ is $\tensor{U} * \tensor{S} * \tensor{V}^{\transpose}$. The singular value thresholding (SVT) operator ($\mathcal{D}_\tau$) is performed on each frontal slice of the f-diagonal tensor $\ftensor{S}$: 
	\begin{equation}
	\begin{gathered}
	\mathcal{D}_\tau (\tensor{X}) \triangleq \tensor{U} * \mathcal{D}_\tau (\tensor{S}) * \tensor{V}^{\transpose}, \ \mathcal{D}_\tau (\tensor{S}) \triangleq \emph{\textsf{ifft}} ( \mathcal{D}_\tau (\ftensor{S}) ), \\
	\mathcal{D}_\tau (\fmat{S}^{(i)}) \triangleq \mathrm{diag} \big( \max\{\sigma_t - \tau, 0\}_{1 \leq t \leq r} \big), \\
	i = 1, 2, \ldots, n_3.
	\end{gathered}		
	\end{equation}
\end{definition}

\section{Tensor truncated nuclear norm} \label{sec:tnnr}

\subsection{Problem formulation}

For tensor $\tensor{X} \Rdim{n_1 \times n_2 \times n_3}$, we define our tensor truncated nuclear norm $\norm{\tensor{X}}_r$ as follows:
\begin{equation} \label{eq:norm_X_r}
\begin{aligned}
\norm{\tensor{X}}_r & \triangleq \norm{\fmat{X}^{(1)}}_r = \sum_{j=r+1}^{\min(n_1, n_2)} \!\! \sigma_j (\fmat{X}^{(1)}) \\
&=  \sum_{j=1}^{\min(n_1, n_2)} \!\! \sigma_j (\fmat{X}^{(1)}) - \sum_{j=1}^{r} \sigma_j (\fmat{X}^{(1)}) .
\end{aligned}
\end{equation}
Combining with Theorem~3.1 in \cite{Hu2013-Accurate}, Theorem~\ref{the:t-SVD}, and Definition~\ref{def:tnn}, we can reformulate \eqref{eq:norm_X_r} as
\begin{align}
\norm{\tensor{X}}_r & \triangleq \norm{\fmat{X}^{(1)}}_* - \max_{\substack{\fmat{A}^{(1)} \fmat{A}^{(1)\transpose} = \bm{I}, \\ \fmat{B}^{(1)} \fmat{B}^{(1)\transpose} = \bm{I}}} \trace{ \fmat{A}^{(1)} \fmat{X}^{(1)} \fmat{B}^{(1)\transpose} } \nonumber \\
&= \norm{\tensor{X}}_* - \max_{\substack{\tensor{A} * \tensor{A}^\transpose = \tensor{I}, \\ \tensor{B} * \tensor{B}^\transpose = \tensor{I}}} \trace{ \tensor{A} * \tensor{X} * \tensor{B}^\transpose } , \label{eq:tnnr_X_r}
\end{align}
where $\tensor{A}$ and $\tensor{B}$ are derived from the t-SVD of $\tensor{X}$. Denote the operator of choosing the initial $r$ columns in the second dimension of $\tensor{U}$ and $\tensor{V}$ (using the Matlab notation) as follows: 
\begin{equation}
\tensor{A} \triangleq \tensor{U}(:,1:r,:)^\transpose,\ \tensor{B} \triangleq \tensor{V}(:,1:r,:)^\transpose. \label{eq:generate_A_B}
\end{equation}

Then \eqref{eq:tnnr_X_r} can be rewritten as the following  problem:
\begin{equation} \label{eq:tnnr_with_max}
\begin{aligned}
\min_{\tensor{X}} \ \ & \norm{\tensor{X}}_* - \max_{\substack{\tensor{A}_\ell * \tensor{A}_\ell^\transpose = \tensor{I}, \\ \tensor{B}_\ell * \tensor{B}_\ell^\transpose = \tensor{I}}} \trace{\tensor{A}_\ell * \tensor{X} * \tensor{B}_\ell^\transpose}  \\
\text{s.t.} \, \ \  & \quad \tensor{X}_{\bm{\Omega}} = \tensor{M}_{\bm{\Omega}} ,
\end{aligned}
\end{equation}
where $\tensor{A} \Rdim{r \times n_1 \times n_3}$ and $\tensor{B} \Rdim{ r \times n_2  \times n_3}$. It is difficult to directly solve \eqref{eq:tnnr_with_max}, so we separate this optimization into two individual steps. First, let $\tensor{X}_1 = \tensor{M}_{\bm{\Omega}}$ as the initial value.  Then in the $\ell$th iteration,  update $\tensor{A}_\ell$ and $\tensor{B}_\ell$ as \eqref{eq:generate_A_B} via the t-SVD (Theorem~\ref{the:t-SVD}). Next, by fixing $\tensor{A}_\ell$ and $\tensor{B}_\ell$, we compute $\tensor{X}_\ell$ from a much simpler problem:
\begin{equation} \label{eq:tnnr_no_max}
\begin{aligned}
\min_{\tensor{X}} \ \ & \norm{\tensor{X}}_* - \trace{\tensor{A}_\ell * \tensor{X} * \tensor{B}_\ell^\transpose}  \\
\text{s.t.} \, \ \  & \quad \tensor{X}_{\bm{\Omega}} = \tensor{M}_{\bm{\Omega}} .  
\end{aligned}
\end{equation}
The detail of solving \eqref{eq:tnnr_no_max} will be presented in the next subsection. By alternately taking two steps above, the optimization will converge to a local minimum of \eqref{eq:tnnr_with_max}. Algorithm~\ref{alg:Tensor-TNNR} summarizes the framework of our method.

\begin{algorithm}[t] \small
	\caption{Tensor truncated nuclear norm for low-rank tensor completion}
	\label{alg:Tensor-TNNR}
	\begin{algorithmic}[1]
		\REQUIRE $\tensor{M}$, the original incomplete data; $\bm{\Omega}$, the index set of  known elements; $\bm{\Omega}^{\text{c}}$, the index set of  unknown elements. \\
		\hspace{-6.15mm} \textbf{Initialization:} $\tensor{X}_1=\tensor{M}_{\bm{\Omega}}$, $\varepsilon = 10^{-3}$, $\ell = 1$, $L = 50$. \\
		\REPEAT
		\STATE \textbf{Step 1:} given $\tensor{X}_\ell \Rdim{n_1 \times n_2 \times n_3}$, calculate 
		\begin{equation}
		[ \, \tensor{U}_\ell,\tensor{S}_\ell,\tensor{V}_\ell \, ]=\text{t-SVD}(\tensor{X}_\ell), \nonumber  \vspace{-0.5ex}
		\end{equation}
		where the orthogonal tensors are
		\begin{align}
		\tensor{U}_\ell \Rdim{n_1 \times n_1 \times n_3}, \ \tensor{V}_\ell \Rdim{n_2 \times n_2 \times n_3}. \nonumber
		\end{align}
		\vspace{-3ex}
		\STATE Compute $\tensor{A}_\ell$ and $\tensor{B}_\ell$ as follows $\left( r \leq \min\{n_1,n_2\} \right)$: 
		\begin{equation}
		\tensor{A}_\ell = \tensor{U}(:,1:r,:)^\transpose,\ \tensor{B}_\ell = \tensor{V}(:,1:r,:)^\transpose. \nonumber 
		\end{equation} 
		\vspace{-3ex}
		\STATE \textbf{Step 2:} solve the optimization problem: 
		\begin{align}
		\tensor{X}_{\ell+1} = &\, \arg\min_{\tensor{X}} \ \norm{\tensor{X}}_* - \trace{\tensor{A}_\ell * \tensor{X} * \tensor{B}_\ell^\transpose} \nonumber \\
		& \quad \ \ \text{s.t.} \quad \ \ \tensor{X}_{\bm{\Omega}} = \tensor{M}_{\bm{\Omega}} . \nonumber 
		\end{align}
		\UNTIL {$\norm{\tensor{X}_{\ell+1} - \tensor{X}_\ell}_\fro \leq \varepsilon$ or $\ell > L$}
		\ENSURE the recovered tensor.		
	\end{algorithmic}
\end{algorithm}

\subsection{Optimization by ADMM}

Because of the convergence guarantee in polynomial time, the ADMM is widely adopted to solve constrained optimization problems, such as Step 2 in Algorithm \ref{alg:Tensor-TNNR}. 
First, we introduce an auxiliary variable $\tensor{W}$ to relax the objective. Then \eqref{eq:tnnr_no_max} can be formulated as 
\begin{equation} \label{eq:step_2_tnnr}
\begin{aligned}
\min_{\tensor{X},\tensor{W}} \ & \norm{\tensor{X}}_* - \trace{\tensor{A}_\ell * \tensor{W} * \tensor{B}_\ell^\transpose} \\
\text{s.t.} \ \ & \, \tensor{X} = \tensor{W}, \ \tensor{W}_{\bm{\Omega}} = \tensor{M}_{\bm{\Omega}} .
\end{aligned}
\end{equation}

The augmented Lagrangian function of \eqref{eq:step_2_tnnr} becomes
\begin{equation}
\begin{aligned}
\mathcal{L}(\tensor{X}, \tensor{W}, \tensor{Y}) &= \norm{\tensor{X}}_* - \trace{\tensor{A}_\ell * \tensor{W} * \tensor{B}_\ell^\transpose} \\
& \quad + \langle \tensor{Y} , \tensor{X} - \tensor{W} \rangle + \frac{\mu}{2} \norm{\tensor{X} - \tensor{W}}_\fro^2,
\end{aligned}
\end{equation}
where $\tensor{Y}$ is the Lagrange multiplier and $\mu > 0$ is the penalty parameter. Let $\tensor{X}_1 = \tensor{M}_{\bm{\Omega}}$, $\tensor{W}_1 = \tensor{X}_1$, and $\tensor{Y}_1 = \tensor{X}_1$ as the initialization. The optimization of \eqref{eq:step_2_tnnr} includes the following three steps:

 \textbf{Step 1}.~Keep $\tensor{W}_k$ and $\tensor{Y}_k$ invariant and update $\tensor{X}_{k+1}$ from $\mathcal{L}(\tensor{X}, \tensor{W}_k, \tensor{Y}_k)$:
\begin{align}
&\tensor{X}_{k+1} = \arg\min_{\tensor{X}} \  \norm{\tensor{X}}_* + \frac{\mu}{2} \norm{\tensor{X} - \tensor{W}_k}_\fro^2 +\langle \tensor{Y}_k , \tensor{X} - \tensor{W}_k \rangle \nonumber \\
& \quad \ \ =  \arg\min_{\tensor{X}} \ \norm{\tensor{X}}_* + \frac{\mu}{2} \Big\| \tensor{X} - \Big( \tensor{W}_k - \frac{1}{\mu} \tensor{Y}_k \Big) \Big\|_\fro^2. \hspace{-1ex} \label{eq:argmin_X}
\end{align}
According to the SVT operator  (Definition \ref{def:SVT}), \eqref{eq:argmin_X} can be solved efficiently by
\begin{equation}
\tensor{X}_{k+1} = \mathcal{D}_{\frac{1}{\mu}} \Big( \tensor{W}_k - \frac{1}{\mu} \tensor{Y}_k \Big) .
\end{equation}

 \textbf{Step 2}.~By fixing $\tensor{X}_{k+1}$ and $\tensor{Y}_k$, we can solve $\tensor{W}$ through
\begin{align}
\tensor{W}_{k+1} &= \arg\min_{\tensor{W}} \ \mathcal{L}(\tensor{X}_{k+1}, \tensor{W}, \tensor{Y}_k) \nonumber \\
&= \arg\min_{\tensor{W}} \ \frac{\mu}{2} \norm{\tensor{X}_{k+1} - \tensor{W}}_\fro^2 - \trace{\tensor{A}_\ell * \tensor{W} * \tensor{B}_\ell^\transpose} \nonumber \\
& \quad + \langle \tensor{Y}_k , \tensor{X}_{k+1} - \tensor{W} \rangle . \label{eq:argmin_W}
\end{align}
Obviously, \eqref{eq:argmin_W} is quadratic with regard to $\tensor{W}$. Therefore, by setting the derivative of \eqref{eq:argmin_W} to zero, we obtain the closed-form solution as follows: 
\begin{equation}
\tensor{W}_{k+1} = \tensor{X}_{k+1} + \frac{1}{\mu} \left(\tensor{A}_\ell^\transpose * \tensor{B}_\ell + \tensor{Y}_k \right) \! .
\end{equation}
In addition, the values of all observed elements should be constant in each iteration, i.e.,
\begin{equation}
\tensor{W}_{k+1} = (\tensor{W}_{k+1})_{\bm{\Omega}^\text{c}} + \tensor{M}_{\bm{\Omega}}.
\end{equation}

 \textbf{Step 3}.~Update $\tensor{Y}_{k+1}$ directly through
\begin{equation}
\tensor{Y}_{k+1} = \tensor{Y}_k + \mu (\tensor{X}_{k+1} - \tensor{W}_{k+1}).
\end{equation}

The concise process is outlined in Algorithm \ref{alg:ADMM}. Since there are merely two variables involved in the convex optimization, the convergence of Algorithm \ref{alg:ADMM} is promised by the alternating direction method.

\begin{algorithm}[t] \small
	\caption{Solving \eqref{eq:step_2_tnnr} by the ADMM}
	\label{alg:ADMM}
	\begin{algorithmic}[1]
		\REQUIRE $\tensor{A}_\ell$, $\tensor{B}_\ell$, $\tensor{M}_{\bm{\Omega}}$, $\mu=5\times 10^{-4}$, $\xi=10^{-4}$, $K = 200$. \\
		\hspace{-6.15mm} \textbf{Initialize:}  $\tensor{X}_1=\tensor{M}_{\bm{\Omega}}$, $\tensor{W}_1 = \tensor{Y}_1 = \tensor{X}_1$, $k=1$. \\
		\REPEAT
		\STATE \textbf{Step 1:} $\tensor{X}_{k+1} = \mathcal{D}_{\frac{1}{\mu}} \! \left( \tensor{W}_k - \frac{1}{\mu} \tensor{Y}_k \right)$. \vspace{0.5ex}
		\STATE \textbf{Step 2:} $\tensor{W}_{k+1} = \tensor{X}_{k+1} + \dfrac{1}{\mu} \left(\tensor{A}_\ell^\transpose * \tensor{B}_\ell + \tensor{Y}_k \right) $. \vspace{0.5ex} \\
		Fix the values of known elements: \vspace{-1ex}
		\begin{equation}
		\tensor{W}_{k+1} = (\tensor{W}_{k+1})_{\bm{\Omega}^\text{c}} + \tensor{M}_{\bm{\Omega}}. \nonumber
		\end{equation}
		\vspace{-3.5ex}		
		\STATE \textbf{Step 3:} $\tensor{Y}_{k+1} = \tensor{Y}_k + \mu \left( \tensor{X}_{k+1} - \tensor{W}_{k+1} \right)$.
		\UNTIL {$\norm{\tensor{X}_{k+1} - \tensor{X}_k}_\fro \leq \xi$ or $k > K$}
		\ENSURE the recovered tensor.		
	\end{algorithmic}
\end{algorithm}

\subsection{Optimization by APGL}

By relaxing the constraint in \eqref{eq:tnnr_no_max}, we rewritten it as
\begin{equation}
\min_{\tensor{X}} \ \norm{\tensor{X}}_* - \trace{\tensor{A}_\ell * \tensor{X} * \tensor{B}_\ell^\transpose} + \frac{\lambda}{2} \norm{\tensor{X}_{\bm{\Omega}} - \tensor{M}_{\bm{\Omega}}}_\fro^2, \label{eq:apgl_objective}
\end{equation}
where $\lambda > 0$ is a penalty parameter.
The APGL method solves the problem in the original form: 
\begin{equation}
\min_{\tensor{X}} \ g(\tensor{X}) + f(\tensor{X}), \label{eq:apgl_original}
\end{equation}
where $g(\cdot)$ is a continuous convex function, $f(\cdot)$ is a convex differentiable function. Instead of directly minimizing \eqref{eq:apgl_original}, the APGL method devises a quadratic approximation of \eqref{eq:apgl_original}, i.e. $Q(\tensor{X},\tensor{Y})$, at a specially chosen point $\tensor{Y}$:
\begin{equation} \label{eq:Q(X,Y)}
\begin{aligned}
Q(\tensor{X}, \tensor{Y}) &= f(\tensor{Y}) + \langle  \nabla f(\tensor{Y}), \tensor{X} - \tensor{Y} \rangle \\
& \quad + \frac{1}{2t} \norm{ \tensor{X} - \tensor{Y} }_\fro^2 + g(\tensor{X}),
\end{aligned}
\end{equation}
where $\nabla f$ is the Fr\'{e}chet derivative of $f(\cdot)$ and $t$ is a scalar. \eqref{eq:Q(X,Y)} can be solved by iteratively updating $\tensor{X}$, $\tensor{Y}$, and $t$. Assume that in the $k$th iteration, $\tensor{X}_{k+1}$ is updated by
\begin{equation}
\begin{aligned}
\tensor{X}_{k+1} &= \arg\min_{\tensor{X}} \, Q(\tensor{X},\tensor{Y}_k) \\
&= \arg\min_{\tensor{X}} \, g(\tensor{X}) + \frac{1}{2t_k} \norm{ \tensor{X} - (\tensor{Y}_k - t_k \nabla f(\tensor{Y}_k)) }_\fro^2.
\end{aligned}
\end{equation}

In accordance with \eqref{eq:apgl_objective}, we define $g(\tensor{X})= \norm{\tensor{X}}_*$ and $f(\tensor{X}) = -\, \trace{\tensor{A}_\ell * \tensor{X} * \tensor{B}_\ell^\transpose} + \frac{\lambda}{2} \norm{\tensor{X}_{\bm{\Omega}} - \tensor{M}_{\bm{\Omega}}}_\fro^2$. By using the SVT operator (Definition~\ref{def:SVT}), we obtain
\begin{align}
\tensor{X}_{k+1} &= \arg\min_{\tensor{X}} \, \norm{\tensor{X}}_* + \frac{1}{2t_k} \norm{ \tensor{X} - (\tensor{Y}_k - t_k \nabla f(\tensor{Y}_k)) }_\fro^2 \nonumber \\
&= \mathcal{D}_{t_k} \Big( \tensor{Y}_k + t_k ( \tensor{A}_\ell^\transpose * \tensor{B}_\ell - \lambda ( \tensor{Y}_k - \tensor{M} )_{\bm{\Omega}} ) \Big).
\end{align}
Then $t_{k+1}$ and $\tensor{Y}_{k+1}$ are updated as the same fashion in \cite{Ji2009-Accelerated}:
\begin{align}
t_{k+1} &= \frac{1 + \sqrt{1 + 4 t_k^2}}{2} , \\
\tensor{Y}_{k+1} &= \tensor{X}_{k+1} + \frac{t_k - 1}{t_{k+1}} \left( \tensor{X}_{k+1} - \tensor{X}_k \right) .
\end{align}

The above procedures are outlined in Algorithm~\ref{alg:APGL}, which holds the convergence rate $O(k^{-2})$ \cite{Beck2009-Iterative}. 

\begin{algorithm}[t]
	\caption{Solving \eqref{eq:step_2_tnnr} by the APGL}
	\label{alg:APGL}
	\begin{algorithmic}[1]
		\REQUIRE $\tensor{A}_\ell$, $\tensor{B}_\ell$, $\tensor{M}_{\bm{\Omega}}$, $\lambda=10^{-2}$, $\xi=10^{-4}$, $K = 200$. \\
		\hspace{-6.85mm} \textbf{Initialize:} $\tensor{X}_1=\tensor{M}_{\bm{\Omega}}$, $\tensor{Y}_1 = \tensor{X}_1$, $t_1 = 1$, $k=1$. \\
		\REPEAT
		\STATE \textbf{Step 1:} \\ $\tensor{X}_{k+1} = \mathcal{D}_{t_k} \! \Big( \tensor{Y}_k + t_k (\tensor{A}_\ell^\transpose * \tensor{B}_\ell - \lambda ( \tensor{Y}_k - \tensor{M} )_{\bm{\Omega}} ) \Big)$.
		\STATE \textbf{Step 2:} $t_{k+1} = \dfrac{1}{2} \left( 1 + \sqrt{1 + 4 t_k^2} \right) $.	\vspace{0.5ex}
		\STATE \textbf{Step 3:} $\tensor{Y}_{k+1} = \tensor{X}_{k+1} + \dfrac{t_k - 1}{t_{k+1}} \left( \tensor{X}_{k+1} - \tensor{X}_k \right)$.
		\UNTIL {$\norm{\tensor{X}_{k+1} - \tensor{X}_k}_\fro \leq \xi$ or $k > K$}
		\ENSURE the recovered tensor.		
	\end{algorithmic}
\end{algorithm}

\section{Experiments} \label{sec:experiment}

In this section, we carry out several experiments to demonstrate the efficacy of our proposed method. The compared approaches are:
\begin{enumerate}
	\item Low-rank matrix completion (LRMC) \cite{Candes2009-Exact};
	\item Matrix completion by TNNR \cite{Hu2013-Accurate};	
	\item Tensor completion by SMNN \cite{Liu2013-Tensor};
	\item Tensor completion by the adaptive tensor nuclear norm (ATNN) minimization \cite{Zhang2018-Nonlocal};
	\item Tensor completion by Tubal-NN \cite{Zhang2014-Novel};
	\item Tensor completion by T-TNN [ours].
\end{enumerate}
The implementation of our algorithm is available online at {\small\url{https://github.com/xueshengke/Tensor-TNNR}}.
All experiments are performed in MATLAB R2015b on Windows 10, with an Intel Core i7 CPU @ 2.60 GHz and 12 GB Memory.
We adjust each parameter of compared methods to be optimal and report the best results. For fair comparisons, each number is averaged over ten individual trials.

In this paper, each algorithm stops if $\norm{\tensor{X}_{k+1}-\tensor{X}_k}_\fro$ is adequately small or the maximum iteration number has reached. Denote $\tensor{X}_{\text{rec}}$ as the final output. Set $\varepsilon = 10^{-3}$ and $L = 50$ for our method. Let $\mu = 5 \times 10^{-4}$ and $\lambda=10^{-2}$ to balance the efficiency and the accuracy of our approach. In practice, the real rank of incomplete data is unknown. Because of the absence of prior knowledge to the number of truncated singular values, $r$ is tested from [1, 30] to manually find the best value in each case.

%In general, the overall reconstruction error (Erec) and the peak signal-to-noise ratio (PSNR) are commonly used metrics to evaluate the performances of different approaches. They are defined as follows:
Generally, the peak signal-to-noise ratio (PSNR) is a widely adopted metric to evaluate the performance of an approach. It is defined as follows:
\begin{align}
\text{MSE} & \triangleq \frac{\norm{(\tensor{X}_\text{rec} - \tensor{M})_{\bm{\Omega}^\text{c}}}_\fro^2}{T}, \\
%\text{MSE} & \triangleq \frac{\text{Erec}^2}{T}, \\
\text{PSNR} & \triangleq 10 \times \log_{10} \! \left( \frac{255^2}{\text{MSE}} \right) \! ,
\end{align}
where $T$ is the total number of missing elements in a tensor, and we presume that the maximum pixel value in $\tensor{X}$ is 255.

\subsection{Video recovery}

\begin{figure}[t]
	\centering
	\subfloat[ 65\% element loss]{\includegraphics[width=0.21\textwidth]{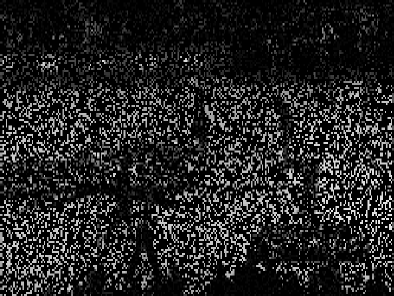}    	\label{subfig:video_incomplete}} \hfil
	\subfloat[PSNR = 17.52, iter = 2084]{\includegraphics[width=0.21\textwidth]{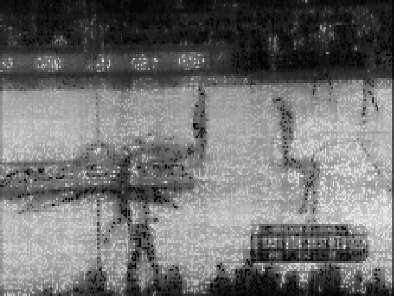}     \label{subfig:video_LRMC}}       \\		\vspace{-1.5ex}	
	\subfloat[PSNR = 18.01, iter = 651 ]{\includegraphics[width=0.21\textwidth]{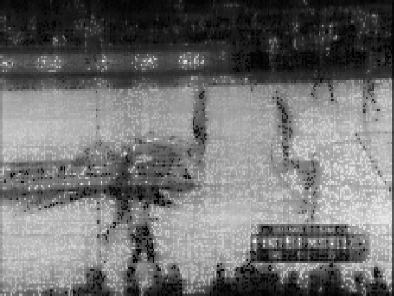}  	 \label{subfig:video_TNNR}}       \hfil
	\subfloat[PSNR = 19.49, iter = 161 ]{\includegraphics[width=0.21\textwidth]{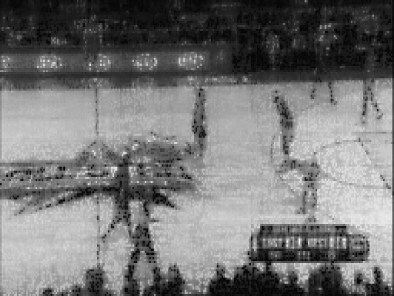}      \label{subfig:video_SMNN}}       \\		\vspace{-1.5ex}
	\subfloat[PSNR = 20.80, iter = 157 ]{\includegraphics[width=0.21\textwidth]{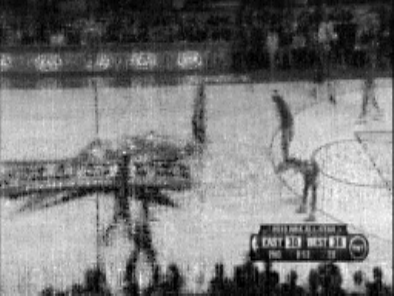}     
	\label{subfig:video_ATNN}}        \hfil
	\subfloat[PSNR = 22.35, iter = 154 ]{\includegraphics[width=0.21\textwidth]{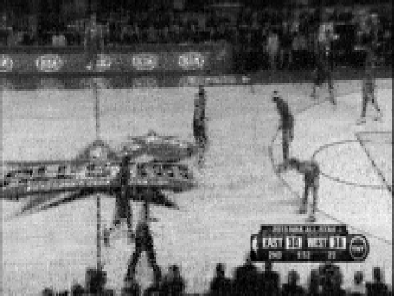}  \label{subfig:video_Tubal-NN}}     \\
	\vspace{-1.5ex}
	\subfloat[PSNR = 24.59, iter = 180 ]{\includegraphics[width=0.21\textwidth]{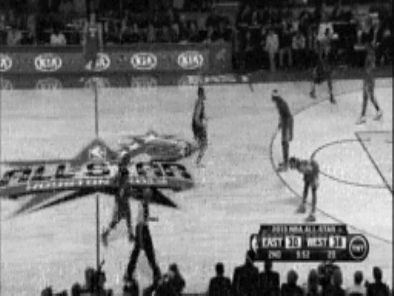}  \label{subfig:video_T-TNN_ADMM}}   \hfil
	\subfloat[PSNR = 24.09, iter = 133 ]{\includegraphics[width=0.21\textwidth]{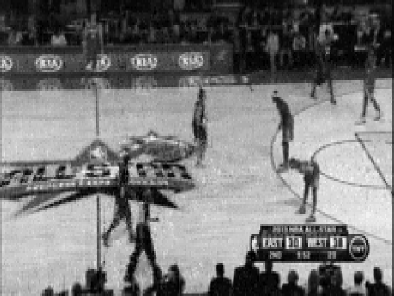}        \label{subfig:video_T-TNN_APGL}}   
	\vspace{-1.5ex}
	\caption{The 20th frame of the basket video reconstructed by seven methods: (a) incomplete frame; (b) LRMC; (c) TNNR; (d) SMNN; (e) ATNN; (f) Tubal-NN; (g) T-TNN ADMM; (h) T-TNN APGL } \label{fig:basket_video}
\end{figure}

We naturally consider videos as 3D tensors, where the first and second dimensions denote space, and the last dimension denotes time. In our experiments, we use a basket video (source: YouTube.com) with size 144\,$\times$\,256\,$\times$\,40, which was captured from a horizontally moving camera in a basketball match. Note that 65$\%$ elements are randomly lost. Fig.~\ref{subfig:video_incomplete} shows the 20th incomplete frame of the basket video. Our T-TNN methods compare to the LRMC, TNNR, SMNN, ATNN, and Tubal-NN methods. The PSNR, iteration number, and the 20th frame of the recovered video are provided in Fig.~\ref{fig:basket_video} to validate the performances of seven approaches.

Obviously, our T-TNN methods (Figs.~\ref{subfig:video_T-TNN_ADMM} and \ref{subfig:video_T-TNN_APGL}) perform much better than other methods. Fig.~\ref{subfig:video_LRMC} shows that the result of LRMC is the worst and requires a large number of iterations, since the matrix completion copes with each frame separately and does not exploit the structural information in the third dimension. Thus, LRMC is not applicable for tensor cases. Beneficial from the truncated nuclear norm, the TNNR (Fig.~\ref{subfig:video_TNNR}) obtains slightly better result than the LRMC and apparently needs less iterations to converge. In tensor cases, Figs.~\ref{subfig:video_SMNN} and \ref{subfig:video_ATNN} reveal similar results, which are the worst compared to the others. This indicates that the SMNN may not be a proper definition for tensor nuclear norm. Figs.~\ref{subfig:video_T-TNN_ADMM} and \ref{subfig:video_T-TNN_APGL} are visually clearer than Fig.~\ref{subfig:video_Tubal-NN},  though the T-TNN ADMM entails more iterations than the Tubal-NN. In addition, the result of T-TNN ADMM (Fig.~\ref{subfig:video_T-TNN_ADMM}) is slightly better than that of T-TNN APGL (Fig.~\ref{subfig:video_T-TNN_APGL}), while the number of iterations of T-TNN APGL is the least than the others. It validates that our T-TNN methods perform best in video recovery.

\subsection{Image recovery}

In real-world applications, numerous images are  corrupted due to  random loss. 
%Hence, tensor completion for image recovery with a subset of data observed is a challenge issue. 
Since a color image has three channels, we hereby deal with it as a 3D tensor rather than separating them in optimization.
%Because a color image has three channels, we hereby regard it as a 3D tensor rather than separating them.

\begin{figure}[t]
	\centering
	\includegraphics[width=0.47\textwidth]{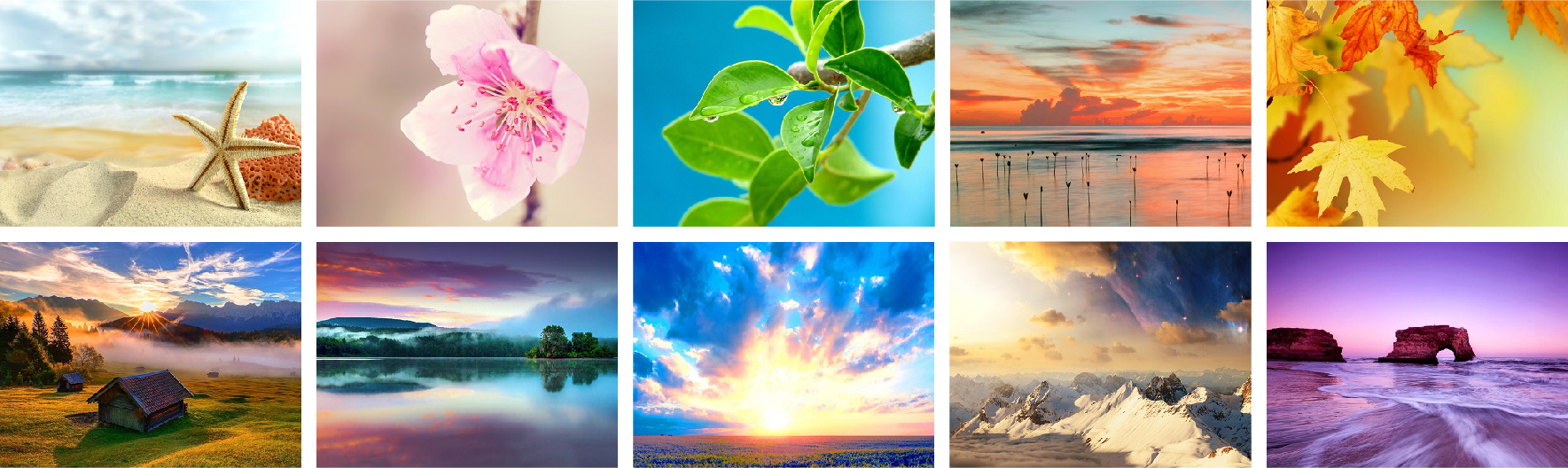}
	\caption{Ten images used in our experiments}
	\label{fig:ten_images}
\end{figure}

\begin{figure}[t]
	\centering
	\subfloat[]{\includegraphics[width=0.21\textwidth]{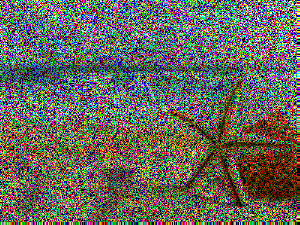} \label{subfig:image_incomplete}} \hfil
	\subfloat[]{\includegraphics[width=0.21\textwidth]{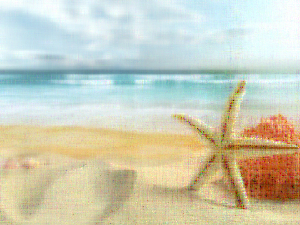}       \label{subfig:image_LRMC}}       \\ \vspace{-1.5ex}	\subfloat[]{\includegraphics[width=0.21\textwidth]{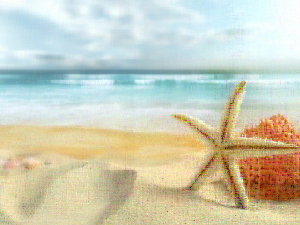}       \label{subfig:image_TNNR}}       \hfil
	\subfloat[]{\includegraphics[width=0.21\textwidth]{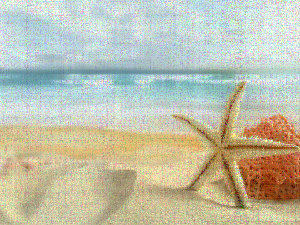}       \label{subfig:image_SMNN}}       \\ \vspace{-1.5ex}
	\subfloat[]{\includegraphics[width=0.21\textwidth]{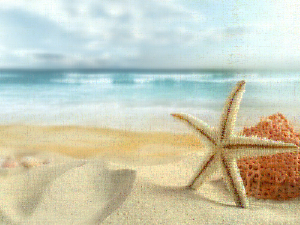}       \label{subfig:image_ATNN}}       \hfil
	\subfloat[]{\includegraphics[width=0.21\textwidth]{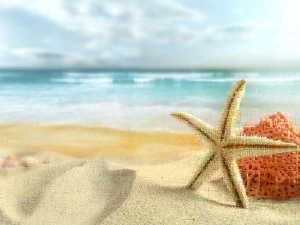}   \label{subfig:image_Tubal-NN}}   \\  \vspace{-1.5ex}
	\subfloat[]{\includegraphics[width=0.21\textwidth]{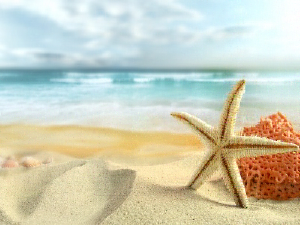} \label{subfig:image_T-TNN_ADMM}} \hfil
	\subfloat[]{\includegraphics[width=0.21\textwidth]{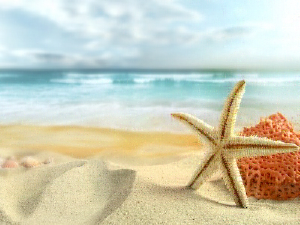} \label{subfig:image_T-TNN_APGL}}     \vspace{-1.5ex}
	\caption{Recovered results of the first image of Fig.~\ref{fig:ten_images} with 50$\%$ random loss by seven methods: (a) incomplete image; (b) LRMC; (c) TNNR; (d) SMNN; (e) ATNN; (f) Tubal-NN; (g) T-TNN ADMM; (h) T-TNN APGL }
	\label{fig:recovered_images}
\end{figure}

\begin{figure}[t]
	\centering
	\includegraphics[scale=0.274]{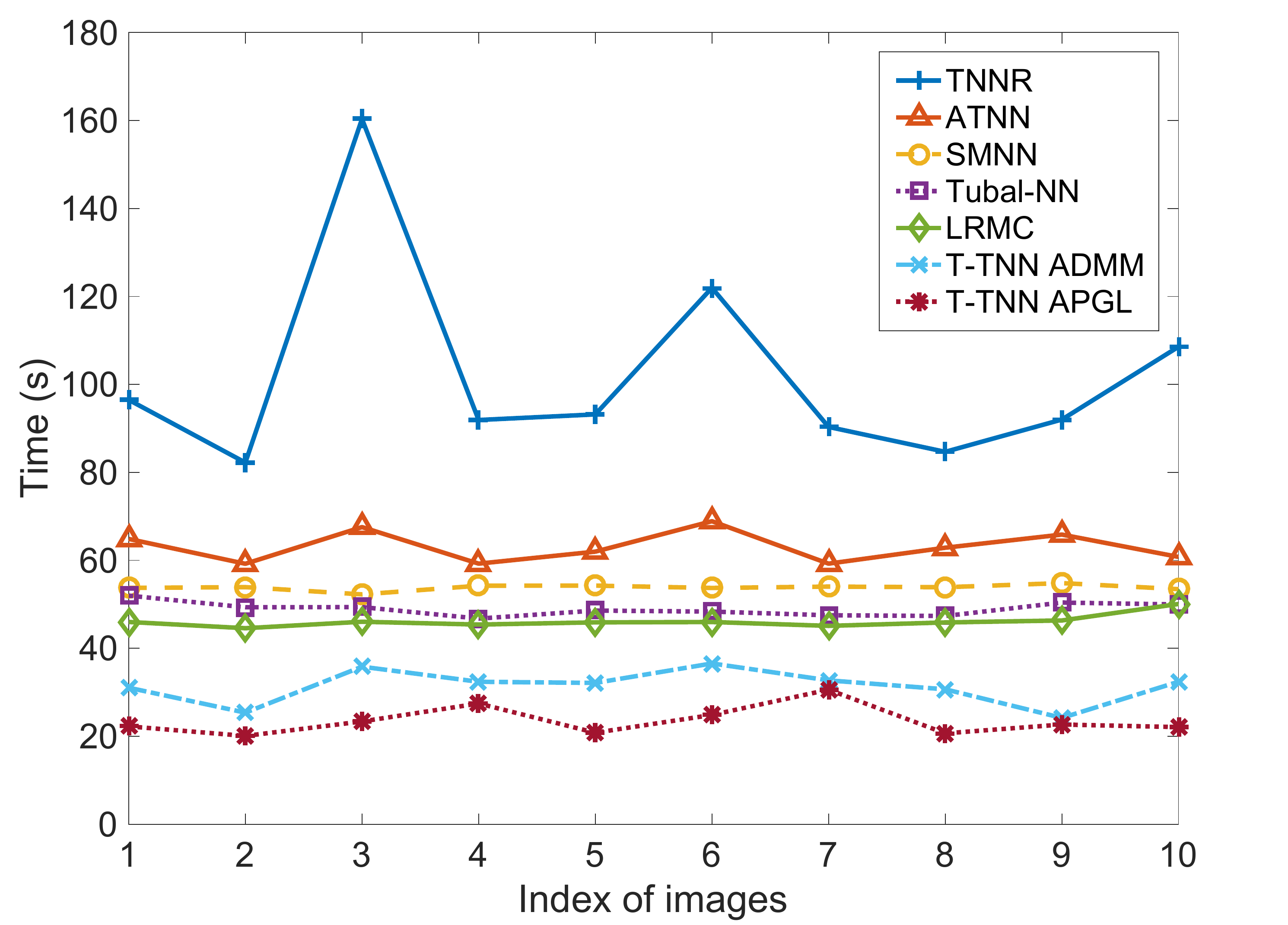}
	\caption{Running time by seven methods on ten images (Fig.~\ref{fig:ten_images})}
	\label{fig:time_comparison}
\end{figure}

In this study, we use ten color images, as shown in Fig.~\ref{fig:ten_images}, all of which are 400\,$\times$\,300 in size. We adopt the PSNR to evaluate the performances of image recovery by different algorithms. Note that 50$\%$ pixels in each image are randomly missing. Fig.~\ref{subfig:image_incomplete} illustrates an example of the incomplete images. 
%Empirically, we set $\mu = 10^{-3}$. 
%Under this configuration, our T-TNN approach compares to the LRMC, TNNR, SMNN, and Tubal-NN methods. The PSNR (iteration), the visualized examples of resulting images, and the running time are adopted as the metrics to assess the performances of seven approaches. They are illustrated in Table~\ref{tab:PSNR_loss}, Fig.~\ref{fig:recovered_images}, and Fig.~\ref{fig:time_comparison}, respectively.
Under this circumstance, our T-TNN approaches compare to the LRMC, TNNR, SMNN, ATNN, and Tubal-NN methods. The PSNR (iteration), the visualized examples of resulting images, and the running time are provided in Table~\ref{tab:PSNR_loss}, Fig.~\ref{fig:recovered_images}, and Fig.~\ref{fig:time_comparison}, respectively.

\begin{table}[t] \footnotesize
	\caption{PSNR of ten recovered images by seven methods with 50$\%$ random element loss (iteration numbers are provided in parentheses)}
	\label{tab:PSNR_loss}
	\centering
	\addtolength{\tabcolsep}{-2pt}
	\begin{tabular}{cccccccc}
		\toprule
		\multirow{2}{*}{No.} & \multirow{2}{*}{LRMC} & \multirow{2}{*}{TNNR} & \multirow{2}{*}{SMNN} & \multirow{2}{*}{ATNN} & \multirow{2}{*}{Tubal-NN} &     T-TNN      &     T-TNN      \\
		                     &                       &                       &                       &                       &                           &      ADMM      &      APGL      \\ \midrule
		 \multirow{2}{*}{1}  &         24.00         &         25.56         &         22.19         &         25.04         &           28.91           & \textbf{29.65} &     29.42      \\
		                     &        (1251)         &         (665)         &         (343)         &         (285)         &           (264)           &     (165)      & \textbf{(125)} \\ \hline
		 \multirow{2}{*}{2}  &         26.51         &         28.79         &         23.19         &         27.43         &           31.45           & \textbf{32.55} &     32.29      \\
		                     &        (1232)         &         (541)         &         (346)         &         (287)         &           (258)           &     (148)      & \textbf{(114)} \\ \hline
		 \multirow{2}{*}{3}  &         20.95         &         24.33         &         22.23         &         24.42         &           26.17           & \textbf{27.67} &     27.42      \\
		                     &        (1274)         &         (878)         &         (341)         &         (291)         &           (262)           &     (178)      & \textbf{(135)} \\ \hline
		 \multirow{2}{*}{4}  &         27.28         &         30.82         &         26.40         &         28.97         &           34.19           & \textbf{35.32} &     35.13      \\
		                     &        (1248)         &         (614)         &         (351)         &         (276)         &           (244)           &     (181)      & \textbf{(142)} \\ \hline
		 \multirow{2}{*}{5}  &         25.91         &         28.96         &         24.95         &         26.34         &           30.29           & \textbf{31.20} &     31.07      \\
		                     &        (1248)         &         (656)         &         (346)         &         (286)         &           (246)           &     (180)      & \textbf{(132)} \\ \hline
		 \multirow{2}{*}{6}  &         22.21         &         23.23         &         22.95         &         23.60         &           25.48           & \textbf{26.24} &     26.13      \\
		                     &        (1251)         &         (813)         &         (339)         &         (277)         &           (241)           &     (212)      & \textbf{(155)} \\ \hline
		 \multirow{2}{*}{7}  &         27.32         &         30.55         &         27.85         &         29.05         &           33.55           & \textbf{34.45} &     34.21      \\
		                     &        (1230)         &         (624)         &         (345)         &         (283)         &           (247)           &     (188)      & \textbf{(127)} \\ \hline
		 \multirow{2}{*}{8}  &         23.85         &         26.04         &         22.80         &         25.68         &           29.02           & \textbf{29.93} &     29.64      \\
		                     &        (1256)         &         (573)         &         (344)         &         (278)         &           (249)           &     (173)      & \textbf{(121)} \\ \hline
		 \multirow{2}{*}{9}  &         22.68         &         24.17         &         22.53         &         24.12         &           27.92           & \textbf{28.98} &     28.60      \\
		                     &        (1261)         &         (639)         &         (347)         &         (283)         &           (260)           &     (140)      & \textbf{(104)} \\ \hline
		\multirow{2}{*}{10}  &         23.52         &         26.60         &         22.40         &         26.28         &           31.59           & \textbf{32.60} &     32.35      \\
		                     &        (1262)         &         (745)         &         (349)         &         (284)         &           (254)           &     (184)      & \textbf{(124)} \\ \bottomrule
	\end{tabular}    
\end{table}

Table~\ref{tab:PSNR_loss} shows the PSNR of seven methods applied on ten images (Fig.~\ref{fig:ten_images}) with 50$\%$ random entries lost. Apparently, the LRMC entails more than 1000 iterations to converge. Based on the truncated nuclear norm, the TNNR noticeably improves the PSNR of recovery on each image and assists convergence, compared with the LRMC. In tensor cases, the SMNN and ATNN perform much worse than the Tubal-NN both in PSNR and iterations, sometimes the SMNN performs worse than the LRMC in PSNR. Note that the results of ATNN are slightly better than the SMNN. This indicates that the SMNN may not be an appropriate definition for tensor completion. The Tubal-NN obtains much higher PSNR and faster convergence than the SMNN and ATNN, which implies that  tensor tubal nuclear norm may be practical for tensor cases. However, our T-TNN ADMM and APGL are both slightly superior in PSNR than the Tubal-NN and obviously converge much faster. In our T-TNN methods, the ADMM holds a bit higher PSNR than the APGL, while the APGL needs less iterations to achieve convergence.

Fig.~\ref{subfig:image_incomplete} presents the first image of Fig.~\ref{fig:ten_images} with 50$\%$ element loss. Figs.~\ref{subfig:image_LRMC}--\ref{subfig:image_T-TNN_APGL} illustrate the recovered results by seven methods, respectively. Apparently, the result of LRMC (Fig.~\ref{subfig:image_LRMC}) contains quite blurry parts. With the help of the truncated nuclear norm, Fig.~\ref{subfig:image_TNNR} is visually much clearer than Fig.~\ref{subfig:image_LRMC}. However, a certain amount of noise still exists. In tensor cases, Fig.~\ref{subfig:image_Tubal-NN} is pretty clearer than Figs.~\ref{subfig:image_SMNN} and \ref{subfig:image_ATNN}, which indicates that the Tubal-NN is much more appropriate than the SMNN. In addition, the result of ATNN (Fig.~\ref{subfig:image_ATNN}) is better than that of SMNN (Fig.~\ref{subfig:image_SMNN}). Moreover, the result of SMNN is even worse than the result of TNNR. The results of our methods (Figs.~\ref{subfig:image_T-TNN_ADMM} and \ref{subfig:image_T-TNN_APGL}) are visually competitive to the result of Tubal-NN (Fig.~\ref{subfig:image_Tubal-NN}), both of which contain only a few outliers, compared with the original image.

Fig.~\ref{fig:time_comparison} presents the running time on ten images by seven methods. Apparently, the TNNR runs much slower than the others and is erratic on different images (from 82.2 s to 160.3 s), since the ADMM intrinsically converges slower and consumes more time on the SVD operator. The ATNN performs relatively stable on these images and much faster than the TNNR, but it is slower than the SMNN. Note that the SMNN performs highly stable and spends about 53.0 s on each image. Similarly, the Tubal-NN and LRMC have nearly identical stability as the SMNN, and they are roughly 3.0 s and 6.0 s in average faster than the SMNN, respectively. In all cases, our T-TNN methods (ADMM and APGL) run quite faster (less than 40 s) than the Tubal-NN and LRMC, though ours are not sufficiently stable. In addition, the APGL is nearly 5.0 s in average faster than the ADMM, since it consumes less iterations. Because our approaches are based on the t-SVD, which has been proved to achieve better convergence. Thus, our T-TNN methods are effective for image recovery and are superior to the compared approaches.

\section{Conclusions and future work} \label{sec:conclusion}
%\section{Conclusions} 

In this paper, we have proposed the tensor truncated nuclear norm for low-rank tensor completion. In detail, we have presented a novel definition of tensor nuclear norm, which is an extension from the standard matrix nuclear norm. The truncated nuclear norm minimization is involved in our approach. We adopt previously proposed tensor singular value decomposition. 
The alternating direction method of multipliers and the accelerated proximal gradient line search method are used to efficiently solve the problem. Hence, the performance of our method is considerably improved. Experimental results show that our approach outperforms the previous methods  in  both recovering videos and images. In addition, the comparison on running time indicates that  our algorithm is further accelerated with the help of the truncated nuclear norm. 

However, there are two deficiencies in our proposed approach. First, the number of truncated singular values $r$ requires manual setting and is sensitive during the optimization process. We consider to make our approach more robust to $r$ during optimization in our subsequent research.
Second, although efficient enough, the ADMM and APGL still entail numerous iterations to solve the sub-problem in our algorithm. Therefore, developing a much faster iterative scheme is a crucial direction in our future work.

\section*{Acknowledgements}

%This work was supported by the Opening Foundation of the State Key Laboratory for Diagnosis and Treatment of Infectious Diseases (No.~2014KF06), the Zhejiang Provincial Natural Science Foundation of China (No.~J20130411), and the National Science and Technology Major Project (No.~2013ZX03005013).

This work is  supported by the Zhejiang Provincial Natural Science Foundation of China (No.~J20130411) and the National Science and Technology Major Project (No.~2013ZX03005013).


%\section*{References}

%%%%%%%%%%%%%%%%%%%%%%%
%% `Elsevier LaTeX' style
%\bibliographystyle{elsarticle-num}
%%%%%%%%%%%%%%%%%%%%%%%

%\bibliography{references}

%%%%%%%%%%%%%%%%%%%%%%%%%%%%%%%%%%%%%%%%

\section*{Appendix} \label{sec:appendix}

%\appendix

\setcounter{equation}{0}
\renewcommand{\theequation}{A.\arabic{equation}}

First, let us recall the definition of the discrete Fourier transform (DFT) of vectors:
\begin{equation}
	\bar{\bm{x}} \triangleq \bm{F} \bm{x},
\end{equation}
where $\bm{F}$ denotes the Fourier matrix, the $(i,j)$th element of which is defined as $\bm{F}_{ij} = w^{(i-1)(j-1)}$, $w = \exp (-\mathrm{j} 2 \uppi / N )$; $\bm{x} = [\bm{x}_1, \bm{x}_2, \ldots, \bm{x}_{N}]^\transpose$ and $\bar{\bm{x}} = [\bar{\bm{x}}_1, \bar{\bm{x}}_2, \ldots, \bar{\bm{x}}_{N}]^\transpose$ are the vectors of time signal and frequency spectrum, respectively. For each element, we have
\begin{equation}
\begin{aligned}
\bar{\bm{x}}_k &= \sum_{n=1}^{N} \bm{x}_n \exp \! \left (-\mathrm{j} \frac{2 \uppi (n-1) (k-1)}{N} \right )  \\
&=  \sum_{n=1}^{N} \bm{x}_n w^{(n-1) (k-1)}, \  k = 1,2,\ldots,N.
\end{aligned}
\end{equation}
If $k = 1$, we obtain $\bar{\bm{x}}_1 = \sum_{n=1}^{N} \bm{x}_n$, i.e., the sum of all elements in $\bm{x}$. Then we consider the DFT in the matrix case and  tensor case.

For tensor $\tensor{X} \Rdim{n_1 \times n_2 \times n_3}$, we have $\ftensor{X} \triangleq \textsf{fft}(\tensor{X},[\,],3)$. Note that the \textsf{fft} function runs along the third dimension. Thus, we compute each element in the Fourier domain as follows: 
\begin{equation}
\begin{gathered}
	\ftensor{X}_{ijk} = \sum_{t=1}^{n_3} \tensor{X}_{ijt} w^{(t-1)(k-1)}, \ w = \exp (-\mathrm{j} 2 \uppi / n_3 ), \\
	i = 1,2,\ldots,n_1, \ j = 1,2,\ldots,n_2, \ k = 1,2,\ldots,n_3.
\end{gathered}
\end{equation}
If $k = 1$, we obtain $\ftensor{X}_{ij1} = \sum_{t=1}^{n_3} \tensor{X}_{ijt}$, i.e., the sum of all elements in $\tensor{X}(i,j,:)$. In the matrix form, we rewrite it as 
\begin{equation}
	\fmat{X}^{(1)} = \sum_{t=1}^{n_3} \bm{X}^{(t)}. \label{eq:proof_x_fourier}
\end{equation}
By using \eqref{eq:proof_x_fourier}, we can efficiently calculate the trace of a tensor in the Fourier domain as follows
\begin{equation}
	\trace{\tensor{X}} = \sum_{t=1}^{n_3} \trace{\bm{X}^{(t)}} = \mathrm{tr} \bigg( \sum_{t=1}^{n_3} \bm{X}^{(t)} \bigg) = \trace{\fmat{X}^{(1)}}.
\end{equation}

Next, we prove the symmetric property \eqref{eq:fft_property}. Based on \eqref{eq:bcirc}, \eqref{eq:unfold}, and Definition \ref{def:tensor_product}, we have
\begin{align}
&\tensor{A} * \tensor{B} = \textsf{fold} (\textsf{bcirc} (\tensor{A}) \cdot \textsf{unfold} (\tensor{B}) ) \nonumber \\
&= \textsf{fold} \! \left ( \! \begin{bmatrix}
	\bm{A}^{(1)}   & \bm{A}^{(n_3)}   & \cdots & \bm{A}^{(2)} \\
	\bm{A}^{(2)}   & \bm{A}^{(1)}     & \cdots & \bm{A}^{(3)} \\
	\vdots         & \vdots           & \ddots & \vdots       \\
	\bm{A}^{(n_3)} & \bm{A}^{(n_3-1)} & \cdots & \bm{A}^{(1)}
\end{bmatrix}
\! \cdot \!
\begin{bmatrix}
	\bm{B}^{(1)} \\
	\bm{B}^{(2)} \\
	\vdots \\
	\bm{B}^{(n_3)} 
\end{bmatrix} \!
\right ) \nonumber \\
&= \textsf{fold} \! \left ( \begin{bmatrix}
	\sum\limits_{i=1}^{1} \bm{A}^{(2-i)} \bm{B}^{(i)} + \sum\limits_{i=2}^{n_3} \bm{A}^{(n_3+2-i)} \bm{B}^{(i)} \\
	\sum\limits_{i=1}^{2} \bm{A}^{(3-i)} \bm{B}^{(i)} + \sum\limits_{i=3}^{n_3} \bm{A}^{(n_3+3-i)} \bm{B}^{(i)} \\
	\vdots \\
	\sum\limits_{i=1}^{n_3} \bm{A}^{(n_3+1-i)} \bm{B}^{(i)}  \\
\end{bmatrix} \right ) . \label{eq:proof_A*B_fold}
\end{align}
Suppose $\tensor{C} = \tensor{A} * \tensor{B}$ and 
\begin{equation}
	\tensor{C} = \textsf{fold} \! \left (
	\begin{bmatrix}
	\bm{C}^{(1)} \\
	\bm{C}^{(2)} \\
	\vdots \\
	\bm{C}^{(n_3)} 
	\end{bmatrix} \right ) , \label{eq:proof_C_fold}
\end{equation} 
then we obtain the following equality by comparing \eqref{eq:proof_A*B_fold} and \eqref{eq:proof_C_fold}:
\begin{equation}
\begin{cases}
	\bm{C}^{(1)} = \sum\limits_{i=1}^{1} \bm{A}^{(2-i)} \bm{B}^{(i)} + \sum\limits_{i=2}^{n_3} \bm{A}^{(n_3+2-i)} \bm{B}^{(i)}, \\
	\bm{C}^{(2)} = \sum\limits_{i=1}^{2} \bm{A}^{(3-i)} \bm{B}^{(i)} + \sum\limits_{i=3}^{n_3} \bm{A}^{(n_3+3-i)} \bm{B}^{(i)}, \\
	\ \ \vdots \qquad \qquad \qquad \vdots \\
	\bm{C}^{(n_3)} = \sum\limits_{i=1}^{n_3} \bm{A}^{(n_3+1-i)} \bm{B}^{(i)}. \\
\end{cases}
\end{equation}
Using the property $\trace{\bm{A} \pm \bm{B}} = \trace{\bm{A}} \pm \trace{\bm{B}}$, we compute 
\begin{align}
	& \trace{ \tensor{A} * \tensor{B} } = \trace{\tensor{C}} = \sum_{i=1}^{n_3} \trace{\bm{C}^{(i)}} \nonumber \\
	& = \mathrm{tr} \bigg( \sum\limits_{i=1}^{1} \bm{A}^{(2-i)} \bm{B}^{(i)} + \sum\limits_{i=2}^{n_3} \bm{A}^{(n_3+2-i)} \bm{B}^{(i)} \bigg) \nonumber \\
	& \quad + \mathrm{tr} \bigg( \sum\limits_{i=1}^{2} \bm{A}^{(3-i)} \bm{B}^{(i)} + \sum\limits_{i=3}^{n_3} \bm{A}^{(n_3+3-i)} \bm{B}^{(i)} \bigg) \nonumber \\
	& \quad + \cdots + \mathrm{tr} \bigg( \sum\limits_{i=1}^{n_3} \bm{A}^{(n_3+1-i)} \bm{B}^{(i)} \bigg) \nonumber \\
	& = \mathrm{tr} \bigg( \sum\limits_{i=1}^{1} \bm{A}^{(2-i)} \bm{B}^{(i)} + \sum\limits_{i=2}^{n_3} \bm{A}^{(n_3+2-i)} \bm{B}^{(i)} \nonumber \\
	& \ \ \quad +        \sum\limits_{i=1}^{2} \bm{A}^{(3-i)} \bm{B}^{(i)} + \sum\limits_{i=3}^{n_3} \bm{A}^{(n_3+3-i)} \bm{B}^{(i)} \nonumber \\
	& \ \ \quad + \cdots + \sum\limits_{i=1}^{n_3} \bm{A}^{(n_3+1-i)} \bm{B}^{(i)} \bigg) \nonumber \\
	& = \mathrm{tr} \bigg( \left ( \bm{A}^{(1)} + \bm{A}^{(2)} + \cdots + \bm{A}^{(n_3)} \right ) \bm{B}^{(1)} \nonumber \\
	& \ \ \quad +        \left ( \bm{A}^{(n_3)} + \bm{A}^{(1)} + \cdots + \bm{A}^{(n_3-1)} \right ) \bm{B}^{(2)} + \cdots \nonumber \\
	& \ \ \quad +        \left ( \bm{A}^{(2)} + \bm{A}^{(3)} + \cdots + \bm{A}^{(1)} \right ) \bm{B}^{(n_3)}  \bigg) \nonumber \\
	& = \mathrm{tr} \bigg( \sum\limits_{i=1}^{n_3} \bm{A}^{(i)} \bm{B}^{(1)} + \sum\limits_{i=1}^{n_3} \bm{A}^{(i)} \bm{B}^{(2)} + \cdots + \sum\limits_{i=1}^{n_3} \bm{A}^{(i)} \bm{B}^{(n_3)}  \bigg) \nonumber \\
	& = \mathrm{tr} \bigg( \sum\limits_{i=1}^{n_3} \bm{A}^{(i)} \left ( \bm{B}^{(1)} + \bm{B}^{(2)} + \cdots + \bm{B}^{(n_3)} \right ) \bigg) \nonumber \\
	& = \mathrm{tr} \bigg( \bigg( \sum\limits_{i=1}^{n_3} \bm{A}^{(i)} \bigg) \bigg( \sum\limits_{i=1}^{n_3} \bm{B}^{(i)} \bigg)\bigg)   .
\end{align}
By using \eqref{eq:proof_x_fourier}, we further obtain
\begin{equation}
	\trace{ \tensor{A} * \tensor{B} } = \trace{ \fmat{A}^{(1)} \fmat{B}^{(1)} }.
\end{equation}
Thus, the proof is accomplished. 

%%%%%%%%%%%%%%%%%%%%%%%%%%%%%%%%%

%\vspace{4ex}
%\begin{window}[0, l, { \mbox{\includegraphics[width=1in,height=1.25in,clip,keepaspectratio]{xsk}} }, {}]
%\noindent \textbf{Shengke Xue}
%received the B.S. degree in communication engineering, from the College of Information Engineering, Zhejiang University of Technology, Hangzhou, China, in 2015. Currently, He is pursuing the Ph.D. degree with the College of Information Science and Electronic Engineering, Zhejiang University, Hangzhou, China. At present, his research interests include pattern recognition, image processing, and machine learning. 
%\end{window}

\end{document}